\documentclass[letterpaper, 10 pt, conference]{ieeeconf}  

\IEEEoverridecommandlockouts                              

\usepackage{graphics} 
\usepackage{epsfig} 
\usepackage{mathptmx} 
\usepackage{times} 
\usepackage{mathtools}
\usepackage{amsmath} 
\usepackage{amssymb}  
\usepackage{color}
\newcommand{\mname}[0]{\textit{3DSGrasp~}}
\newcommand{\mnameNoSpace}[0]{\textit{3DSGrasp}}

\usepackage{boldline}

\newcommand\coolover[2]{\mathrlap{\smash{\overbrace{\phantom{%
    \begin{matrix} #2 \end{matrix}}}^{\mbox{$#1$}}}}#2}

\title{3DSGrasp: 3D Shape-Completion for Robotic Grasp}

\author{Seyed S. Mohammadi$^{2,3}$ \and Nuno F. Duarte$^{1}$ \and Dimitris Dimou$^{1}$ \and Yiming Wang$^{3,4}$ \and Matteo Taiana$^{3}$ \and Pietro Morerio$^{3}$ \and Atabak Dehban$^{1}$ \and Plinio Moreno$^{1}$ \and Alexandre Bernardino$^{1}$ \and Alessio {Del Bue}$^{3}$ \and José Santos-Victor$^{1}$ 
\thanks{*This work has partially received funding from the European Union’s Horizon 2020 research and innovation programme under grant agreement No 964854; the FCT funding to the ISR/LARSyS Associated Laboratory UID/EEA/50009/2020 and LA/P/0083/2020 
N. F. Duarte is supported by FCT-IST fellowship grant PD/BD/135116/2017.}
\thanks{$^{1}$Vislab, Institute for Systems and Robotics|Lisboa, Instituto Superior T\'{e}cnico, Universidade de Lisboa, Portugal. Email:{\tt\small$\{$nferreiraduarte, plinio, alex, jasv$\}$@isr.tecnico.ulisboa.pt}} %
\thanks{$^{2}$Department of Marine, Electrical, Electronic and Telecommunications Engineering, University of Genoa, Italy.}
\thanks{$^{3}$Pattern Analysis \& Computer Vision (PAVIS), Istituto Italiano di Tecnologia (IIT), Genoa, Italy. Email:{\tt\small$\{$seyed.mohammadi,  yiming.Wang, matteo.taiana, pietro.morerio, alessio.delbue$\}$@iit.it}}
\thanks{$^{4}$Deep Visual Learning (DVL), Fondazione Bruno Kessler, Trento, Italy.}}

\begin{document}

\maketitle
\thispagestyle{empty}
\pagestyle{empty}


\begin{abstract}

Real-world robotic grasping can be done robustly if a complete 3D Point Cloud Data (PCD) of an object is available.  
However, in practice, PCDs are often incomplete when objects are viewed from few and sparse viewpoints before the grasping action, leading to the generation of wrong or inaccurate grasp poses. 
We propose a novel grasping strategy, named \mnameNoSpace, that predicts the missing geometry from the partial PCD to produce reliable grasp poses. Our proposed PCD completion network is a Transformer-based encoder-decoder network with an Offset-Attention layer. Our network is inherently invariant to the object pose and point's permutation, which generates PCDs that are geometrically consistent and completed properly. Experiments on a wide range of partial PCD show that \mname outperforms the best state-of-the-art method on PCD completion tasks and largely improves the grasping success rate in real-world scenarios. The code and dataset will be made available upon acceptance.

\end{abstract}

\section{INTRODUCTION}

Robotic grasping has recently gained increasing  attention because of its essential role in many real-world applications, such as domestic and collaborative robotics. The seminal work of Pas et al. \cite{ten2018using} uses 3D Point Cloud Data (PCD) to generate grasp poses directly on the available  3D object structure. However, in real practical scenarios we often have to rely on  incomplete geometric information acquired from single or few viewpoints, which leads to drastic reduction of grasping success rate.

Researchers bypassed this problem by acquiring complete 3D object scans \cite{lu2022online} but this requires a feasible camera path
around the object, which is time consuming to obtain and not always feasible.
Another strategy is to place additional sensors around the object of interest \cite{lin2020robotic}, but this is not cost-effective and it requires careful calibration.


Instead, this paper aims at improving single-view grasping by predicting the missing geometrical structure from a partial PCD. 3D shape completion is an inherently ambiguous problem but recent learning-based approaches have provided  encouraging results on different classes of objects. 
Initial shape completion solutions \cite{varley2017shape, lundell2019robust} converted the 3D point cloud to a voxel grid with the rendering of additional data that increases processing time and memory requirements. More efficient networks \cite{yuan2018pcn, liu2020morphing} were inspired by the PointNet \cite{qi2017pointnet} architecture that directly processes unordered PCDs. 
However, most of these methods have been evaluated on synthetic, noise-free datasets, far from real-world  scenarios. Differently, this work proposes a new model for 3D point completion that can operate in a realistic scenario for robot grasping with arbitrary object classes. 
Our method adopts a transformer-based network \cite{dosovitskiy2020image} and it proposes a modification of an Offset-Attention layer \cite{wang2021poat, guo2021pct} with the introduction of skip connections that is able to complete the partial PCD as extracted from just a single depth camera frame. By completing the point cloud, the computation of the grasp poses can leverage the additional information of a full PCD.

\begin{figure}[t!]
  \centering
  \includegraphics[width=0.5\textwidth]{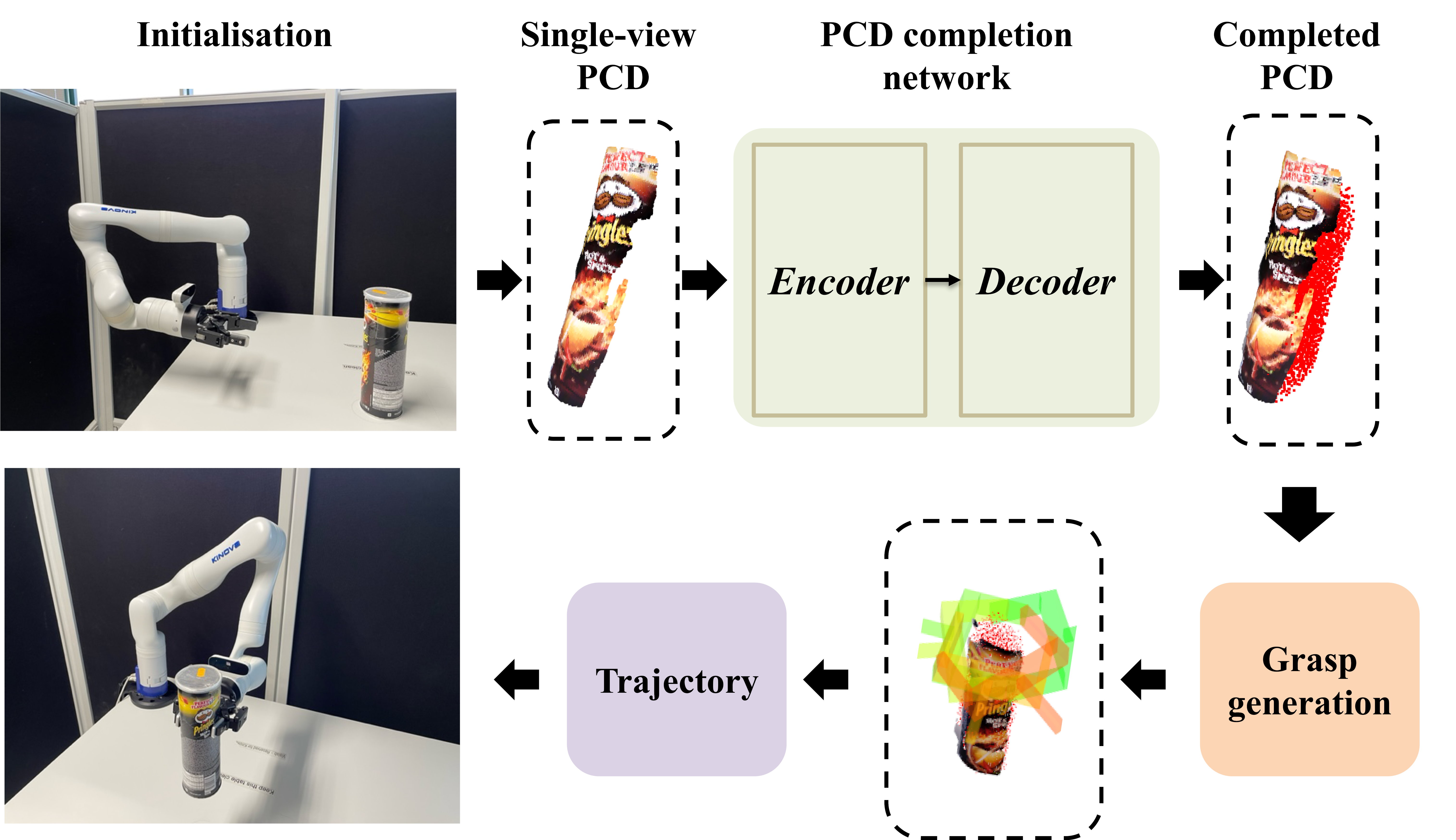}
  \vspace{-0.25cm}
  \caption{Overall pipeline of the proposed 3D robotic grasping strategy. We first capture a partial PCD from a single view of the object using a depth sensor located on the Kinova robotic arm. We then 
  feed the single-view PCD to the completion network and produce a completed PCD. Finally, we generate the grasp pose and execute the grasp with feasible trajectory for the robot.}
  \label{fig:first_ARC}
\end{figure}

Our proposed grasping pipeline is shown in Figure \ref{fig:first_ARC}. With the calibrated camera equipped on the robotic arm, we first acquire the PCD and segment the background information using PointNet++ \cite{qi2017pointnet++}. The segmented partial PCD of the object is then normalised, i.e. scaled and centered, and fed to the PCD completion network to predict the missing geometry of the object. We then map back the predicted point cloud in the real-world scene reference in order to merge the predicted missing PCD with the observed partial input. Furthermore, we generate the grasp pose on the top of virtually completed point cloud using the method proposed in Grasp Pose Detection in Point Clouds (GPD) \cite{ten2017grasp}. Finally, we utilise Moveit! \cite{coleman2014reducing} to plan the arm trajectory that moves the gripper to the pose estimated by GPD. 

We first evaluate our PCD completion method on a PCD completion benchmark dataset \cite{varley2017shape} that has been generated on  the top of  YCB dataset \cite{calli2015ycb}, by training all the state-of-the-art methods (from scratch) using the same dataset (and split), and outperform the reconstruction error of the best state-of-the-art methods.  Then, we test the proposed grasping pipeline in a real scenario using a Kinova arm, our completion network, and GPD. Our method provides accurate completions for successful grasp poses, which enclose the self-occluded parts of the object. Thus, the set of promising grasp hypothesis is larger, which improves the overall success rate score.



To summarise, these are our main contributions:

\begin{itemize}
\item  We propose a novel partial PCD completion network based on the Offset-Attention encoder-decoder Transformer, that achieves state-of-the-art PCD completion performance when evaluated on the partial version of the YCB dataset proposed in \cite{varley2017shape}.



\item We integrate and test our grasping pipeline with a Kinova arm, showing a significant improvement in robotic grasping success rate. 

\item We present extensive ablation studies on the architecture of our proposed completion network to best justify our design choices. 

\end{itemize}

\section{RELATED WORK}
We mainly cover related works addressing shape completion  with 3D data and robotic grasping.
\\
\noindent \textbf{3D shape completion.} In environments where objects are not placed on top of others, such as cupboards and shelves, object shape completion can provide additional grasp poses that augment the selection range. 

Given the incomplete partial 3D data as the input, the aim is to predict an approximation of the complete shape. 3D shape completion methods can be categorised into geometric and data-driven approaches \cite{han2017high}. Geometry-based methods \cite{kazhdan2006poisson,de2015efficient} assume the presence of shape priors, such as geometric primitives, symmetry and structural repetition \cite{berger2014state}. However, the application of these priors may lead to less accurate reconstructions for large-scale datasets and real-world 3D data. Data-driven (i.e. learning-based) approaches rely on deep neural networks that discover the shape completion priors from the data both at local and global point cloud level \cite{gurumurthy2019high, han2017high}. 
 
In earlier works, the irregular 3D data (i.e., raw point cloud) is converted to a regular data representation (i.e.,voxel grid),
where 3D CNNs applied on voxelized inputs have been widely adopted for the pure 3D shape completion task \cite{han2017high}  and for shape completion for improving grasp estimation \cite{lundell2019robust, varley2017shape}. However, the cost of memory usage and computational time for such methods is very large \cite{qi2017pointnet}.

Instead, PCN \cite{yuan2018pcn} directly uses raw PCD for shape completion tasks, and it is based on an encoder-decoder architecture. The encoder is a PointNet-based backbone network that provides global features. The decoder has two stages, one that estimates a coarse point cloud by applying an MLP. After that, FoldingNet \cite{yang2018foldingnet} is used to generate the detailed and completed point cloud. 
Following PCN, a range of learning-based methods for pure 3D shape completion tasks from PCD were proposed \cite{liu2020morphing, groueix2018papier, pan2021multi, yu2021PoinTr} to improve the resolution and robustness of the reconstructed PCD, while others~\cite{yang2021robotic,PointSDFgrasping2020, chen2022improving}, were proposed for improving the performance of grasp success rate by directly processing 3D PCD for completing the shape of the object.

PoinTr \cite{yu2021PoinTr} was the first PCD completion system to adopt the Transformer architecture \cite{dosovitskiy2020image}, leading to a significant improvement in performance.  
Later, \cite{chen2022improving} introduced a transformer-based network for object completion 
that consists of an encoder-decoder architecture, where the encoder is a conventional Multi-Head Self-Attention module, and the decoder is based on the AtlasNet \cite{groueix1802atlasnet}. Although the authors improve the reconstruction result of the GRnet \cite{xie2020grnet} network that use 3D grids to regularize unordered PCD  for point cloud completion, they do not compare their results with PoinTr \cite{yu2021PoinTr}, which consistently outperforms GRnet. In addition, the alignment between the partial point cloud and the reconstructed one requires a 6D pose estimation module.
In contrast, our method accurately aligns the observed point cloud with the reconstructed one.
Additionally, according to our experiments, we improve the reconstruction results compared to the state-of-the-art and provide more promising grasp poses.
\\
\noindent \textbf{Vision based Robotic Grasping.} Robotic Grasping aims to find the optimal pose of the robot's end-effector that leads to a successful grasp of an object. In one way, model-based methods consider contact points and exerted forces to select the grasps that provide more stability, but the evaluation is usually in simulation, which suffers from a large reality gap~\cite{du2021vision}. In the other, data-driven approaches aim to map directly perceptual input such as RGB \cite{lenz2015deep,7139361} and RGB-D images \cite{fang2020graspnet,graspnet4,sundermeyer2021contact}, to the grasp success. Recent methods take advantage of model-based and data-driven approaches by generating data samples and labels from simulations using domain randomization \cite{fang2020graspnet,graspnet4,sundermeyer2021contact}.

Current approaches are able to map 6DoF pose candidates to point clouds \cite{mousavian20196,ten2017grasp,sundermeyer2021contact,zhao2021regnet,qi2017pointnet++}, addressing successful grasping in cluttered scenarios. From the perceptual point of view, segmentation of the objects is very challenging, so these approaches start by sampling grasp poses, followed by grasp pose score computation and finally a refinement pose procedure. Amongst the 6DoF grasping approaches, we select Grasp Pose Detection in Point Clouds (GPD) \cite{ten2017grasp} to be used in our system, due to the computational efficiency of the grasp sampling and score computation \cite{zhao2021regnet}. The main steps of GPD are: (i) heuristics-based grasp candidate sampling and (ii) binary classification of candidates by a CNN. A detailed description of GPD is in Section \ref{sec:method:Grasp pose generation and evaluation}.

\begin{figure*}[t!]
  \centering
   \includegraphics[width=1\textwidth]{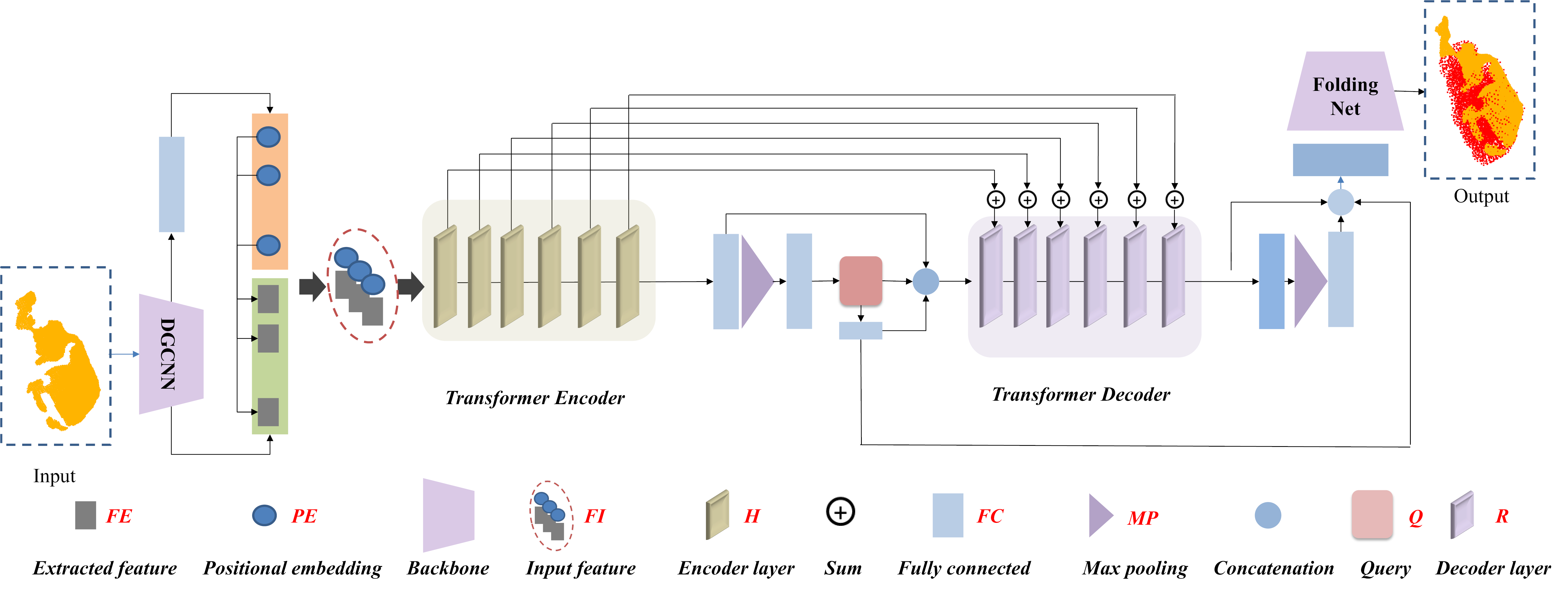}
  \vspace{-0.25cm}
  \caption{Architecture of our point cloud completion network. Given the partial PCD as the input, we first apply FPS to the subset of the points representing the center point $CR$ of each local region $LR$. Then we use $KNN$ to gather the points around each $CR$ and send them to DGCNN to extract embedding feature $FE$. We then send the $CR$ to a $FC$ layer to learn the Positional Embedding $PE$.  Furthermore, we concatenate $PE$ and with the corresponding   $FE$ to be the input of the Transformer. As the output we predict the shape feature for a missing PCD $PM$ and fed to the  FoldingNet to generate high resolution PCD, then we merge the input PCD with the predicted output PCD to shape the completed PCD $PC$.}
  
  \label{fig:ARC}
\end{figure*}


\section{APPROACH}
We assembled the setup as a Kinova robotic arm equipped with a RealSense depth camera and Robotiq gripper; and an object $O$ to be grasped. We then utilise the depth camera to capture a depth image from a single viewpoint of the scene. Furthermore, we convert the depth image to PCD using camera parameters. The reconstructed PCD contains only the visible part of the object from the camera's point of view (i.e partial 3D scan).

Given a partial 3D scan, containing background information and a colourless partial PCD, we first segment the partial PCD $PP$, $PP = \left\{PP^i \mid PP^i \in \mathbb{R}^3, i=1 \dots N\right\}_{N{= 2048}}$, using the PCD segmentation network presented in \cite{qi2017pointnet++}.
Then, we use our proposed completion network for processing $PP$ to predict the missing PCD $PM$, $PM = \left\{PM^i \mid PM^i \in \mathbb{R}^3, i=1 \dots M\right\}_{M{= 6144}}$, representing the missing point cloud of the complete shape. Finally we map back the predicted missing PCD to the real scene and  merge it with  the partial PCD.
Furthermore, we generate a grasp candidate $GK$ on the \emph{completed} PCD using the Grasp Pose Detection (GPD) network \cite{ten2017grasp}, which outputs a set of grasp poses $\mathbf\{G_k\}$, $GK = \{{GK}_1, {GK}_2, ... {GK}_V\}_{{V = 5}}$ with their corresponding classification scores $CS$. Lastly, the grasp with the best classification score $GK_{BCS}$ that is considered feasible by MoveIt!~\cite{coleman2014reducing}, is executed on a real robot.

In Section \ref{sec:method:alignment}, we introduce a dataset pre-processing step and address the PCD alignment problem for PCD completion task. Section \ref{sec:method:completion} describes the proposed point cloud completion network in detail and the defined loss functions used for training the network. Finally, Section \ref{sec:method:Grasp pose generation and evaluation} describes the grasp pose generation and evaluation network.

\subsection{PCD alignment pre-processing }
\label{sec:method:alignment}

Data normalisation is a primary stage for improving the generalisation of deep models on the learning process~\cite{bengio2012practical}. 
%
However, standard PCD completion approaches use a data normalisation that is not applicable to grasping problems. The centroid of each PCD in training is given by the centroid of the completed (full PCD) object, either from CAD model~\cite{mohammadi2021pointview} or the GT PCD~\cite{yuan2018pcn}. This is not an issue in general, as PCD completion protocols during testing provide the partial shape aligned with the centroid of GT. 
Differently, in  a real testing scenario, the shape centroid can only be computed from partial PCDs and thus being different from the GT one. For this reason, the pre-processing of PCD in training has to take into account that the centroid available is only related to the partial PCD. Otherwise the completed PCD would be misaligned as shown in the ablation studies in Section \ref{exp:abl_arc_}.

In this work, we have proposed a simple but an effective technique to solve this problem without using GT information.
Given the partial PCD $\{PP\}$, we first calculate the translational offset vector $\{T_p \in \mathbb{R}^3\}$, where $T_p = \frac{1}{N}\sum_{i=1}^N PP^i$.  We then calculate the centered PCD $\{PP\}^c$ as; $ {PP}^c  = PP - T_p$. Furthermore, we normalise  the scale $\{S_P \in \mathbb{R}\}$ as; $S_P  = \max_i \|PP^i-T_p\|_2$, where $\|\cdot\|_2$ is norm-2 and the final normalise PCD $\{{PP}^n\}$ will define as; ${PP}^n  = {PP}^c/S_P$.

$T_p$ and $S_P$ are the normalisation parameters calculated from the partial point cloud.
However, to avoid the misalignment phenomenon,
instead of separately  calculating  the offset and the scaling for ground-truth point cloud $PGT$, we simply apply the same parameters~(\textit{i.e.}, $T_p$ , $S_p$ ) achieved by the normalisation  of  $PP$ on $P_{gt}$ such that: ${PGT}^c = PGT - T_p$  and ${PGT}^n = {PGT}^c/S_p$. In this way, we precisely align the ground-truth PCD with the partial one, and also we consider the partial PCD as a reference PCD which is the application for a real-world scenario. After normalising the dataset, we apply $FPS$ to sample $2048$ points for a partial PCD and $8192$ points for the Ground-Truth PCD.
%

\subsection{Point cloud completion network}
\label{sec:method:completion}
This section illustrates in detail how the proposed Transformer completion network predicts the missing geometry of the 3D data. The architecture is inspired from \cite{yu2021PoinTr}, but using an Offset-Attention \cite{guo2021pct} instead of the usual Self-Attention encoder-decoder block, which was shown be more suitable to process PCD given its intrinsic invariance to rigid transformation. Moreover, we propose Skip-Connections among the layer of the encoder and decoder for the better generalisation of the network.
The network is composed of three main blocks: The PCD embedding, the Transformer block consisting of the Offset-Attention encoder-decoder layer, and the block that generates the PCD for the missing part.

\subsubsection{\textbf{Point cloud embedding}}

The Transformer architecture requires an ordered sequence of vectors (\textit{e.g.} like words in a sentence). However, PCD is invariant to permutations, (\textit{i.e.}, by changing the point sequence order there should be no difference in the description of the shape of the object).

To address this property of PCD, in this work, we follow the pipeline  as in
\cite{yu2021PoinTr}. We divide partial PCD $PP$  into the set of Local Regions $LR$, $LR = \{{LR}_1, {LR}_2, ... {LR}_R\}_{R{= 128}}$ by applying $Farthest Point Sampling$ ($FPS$) \cite{qi2017pointnet++} and then we represent the centroid as $CR$, $\left\{CR^i \mid CR^i \in \mathbb{R}^3, i=1 \dots B\right\}_{B{= 128}}$,  of each  $LR$. We then apply $KNN$ \cite{dgcnn} to find  the points around each $CR$. Furthermore, we feed the points in each $LR$ into the PCD-backbone network \cite{dgcnn} (DGCNN) to compute the embedding feature $FE$, $FE = \{ FE_1, FE_2, ..., FE_ B\}_{B{= 128}}$.  We also feed each $CR$ to the  $fully-connected$~$(FC)$ layer  to extract the positional embedding $PE$ (i.e.,  describe the location of each subset of the points in each $LR$  \cite{dosovitskiy2020image}) for each $FE$. Finally, we concatenate the $PE$ with the corresponding $FE$ to be the input $FI$, $FI = \{{FI}_1, {FI}_2, ... {FI}_J\}_{J{= 128}}$ of the Offset-attention Transformer encoder network.     

\subsubsection{\textbf{Transformer architecture}}

\begin{figure}[t!]
    \centering
    \includegraphics[width=0.5\textwidth]{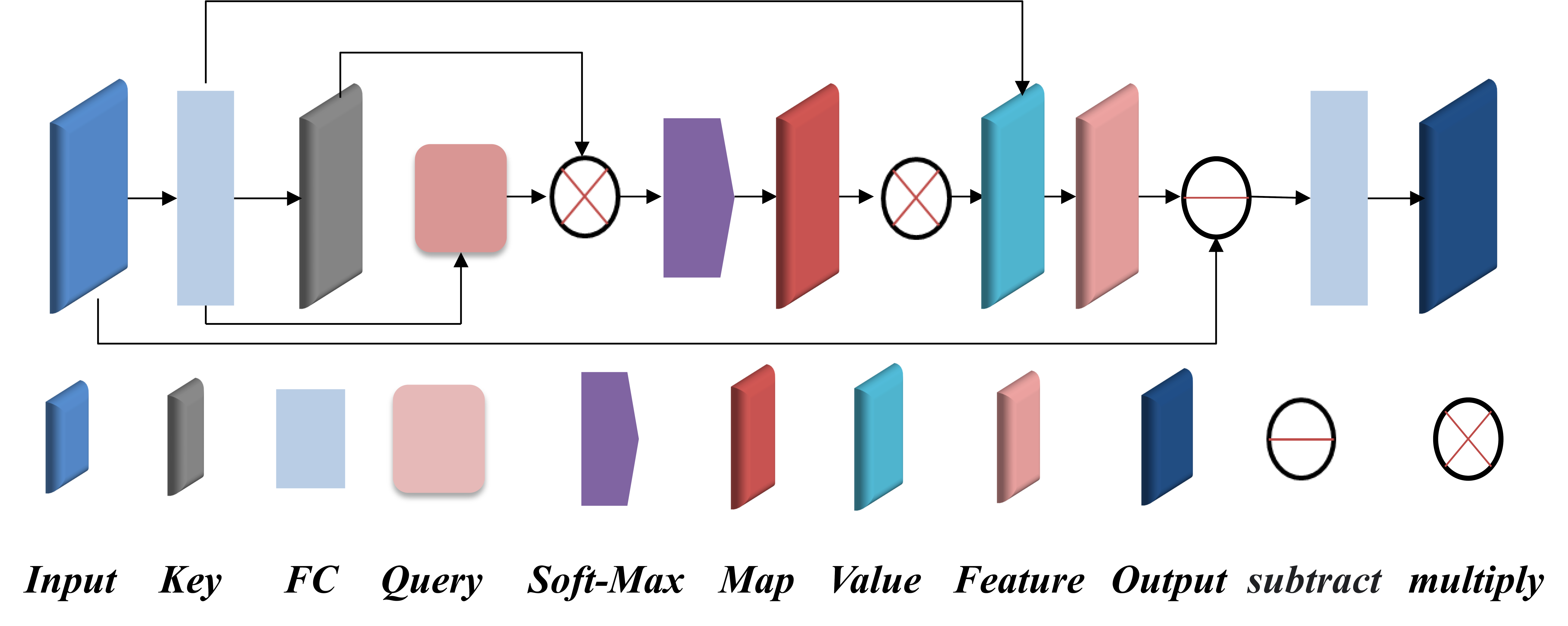}
     \vspace{-0.25cm}
    \caption{The Offset-Attention layer measure the difference between the Attention and the input feature.}
    \label{fig:atten}
\end{figure}

We propose to use a multi-head Offset-Attention encoder-decoder Transformer layer \cite{guo2021pct, wang2021poat} for PCD completion task, since Offset-Attention layers have been shown to be advantageous over the usual self-attention layer on point cloud segmentation and classification. This is specially important in robotic grasping contexts where the relative pose between the object and end-effector is arbitrary. The real-world point cloud completion task must be independent from the initial pose of the object as the camera can see the object from different positions. By using the Offset-Attention layer, we take advantage of its invariance to rigid transformations, resulting in a more robust object completion. Fig. \ref{fig:atten} shows the architecture of the Offset-Attention layer, where the offset is calculated by measuring the difference between the input features $FI$ and Self-attention features $SA$, $SA = \{ SA_1, SA_2, ... SA_J\}$  by subtracting one from the other $FI_J - SA_J$. 

Given a sequence of the input features $FI$, we formulate the encoder as: $AE = E (FI)$, 
where $E$ is the encoder and $AE = \{{AE}_1, {AE}_2, ... {AE}_w\}_{W{= 1024}}$ is the output feature vector of the encoder.
The Offset-Attention in the encoder layer first updates the input features $FI$. Then, we feed the output of the encoder to the $FC$ layer, followed by a $Max-Pooling$~$(MP)$ operation. 
Moreover, to force the encoder to learn and generalise better about the global complete shape information, we predict the sparse PCD $PS$, $PS = \left\{PS^i \mid PS^i \in \mathbb{R}^3, i=1 \dots S\right\}_{S{= 192}}$, where $PS$ is the predicted PCD, representing the complete shape of the object with lower number of the points. We predict $PS$ by passing the generated feature vector $AE$ (i.e., the output of the encoder) to the queries $Q$ layer that  containing  $FC$. Then we reshape the output of the $FC$ layer to $S \times 3$ to create $PS$.




On the other hand, the Offset-Attention decoder layer $D$ shares the exact architecture of the encoder network except having cross-attention mechanisms \cite{lin2022cat}. We formulate the  decoder architecture as $AD =  D (Q, H)$, where $Q = \{Q_1, Q_2, ... Q_X\}_{X{= 192}}$ is a set of queries and $AD = \{AD_1, AD_2, ... AD_Y\}_{Y{= 512}}$  are the predicted output features representing the feature vector of the missing point cloud.

\begin{table*}[t]

\caption{COMPARISON OF $L2$ $ CD$   LOSS IN DIFFERENT POINT CLOUD COMPLETION MODELS ON YCB DATASET. WE REPORT THE RESULT OF 14 SEEN CATEGORIES AND 4 UNSEEN CATEGORIES.}
\vspace{-0.25cm}
\centering
\resizebox{1\textwidth}{!}{%
\begin{tabular}{|l|c|c|c|c|c|c|c|c|c|c|c|c|c|c!{\vrule width2.0pt}c|c|c|c|}
\hline
 Method & Avg& Drill box & Mini Soccer ball & Tomato Soup & Cleanser & Comet Bleach & Box of Sugar & Mustard & Lemon & Morton Salt & Pringles & Pitcher& Sponge& Cup & Block  & Cracker Box & Banana & Stack Blocks \\ 
\hline
TopNet \cite{tchapmi2019topnet} &2.51&2.18 & 2.24 &1.98 & 1.84&  1.84& 1.79&2.01&2.51& 1.87& 1.70&  1.99&2.24& 2.84&3.32&3.56 &3.90&7.52\\ 
FoldingNet \cite{yang2018foldingnet} &2.28&2.01& 2.19&1.81 & 1.66& 1.59& 1.48& 1.87& 2.29& 1.63&1.55 & 1.82& 2.18&2.53 &3.10 &3.14&3.33 &7.01\\ 
PCN \cite{yuan2018pcn}  &2.07&1.86 &1.95 & 1.59&1.48 &1.41 &1.62 & 1.98& 1.42&1.50 &1.46 & 1.69&2.00 & 2.05&2.89 &2.93&2.99&6.53\\ 
MSN \cite{liu2020morphing} &1.98& 1.81  &2.0 & 1.49& 1.39& 1.44&1.51 & 1.88& 1.25& 1.38&1.52 & 1.63& 1.84& 1.91&2.92& 2.65&2.78&6.41\\  

PoinTr \cite{yu2021PoinTr} &1.15& 0.83 & \textbf{0.95}& 0.99 & 0.83& 0.68& 0.64& 0.79&\textbf{0.91} & 0.83&0.61 & 0.72 &0.87& 0.98&1.46&1.32&1.5&5.91\\
\hline


\mname  &\textbf{0.92}& \textbf{0.64}& 1.00 &\textbf{0.81} &\textbf{0.52} &\textbf{0.49} & \textbf{0.48} & \textbf{0.53}&  \textbf{0.99} & \textbf{0.65}&  \textbf{0.40} &\textbf{0.68} &\textbf{0.51}& \textbf{0.70} & \textbf{1.18}& \textbf{1.15} & \textbf{0.98}&\textbf{4.89}\\

\hline
\end{tabular}}
\label{exp:tab:comp}
\end{table*}

\begin{table}[t!]
\centering
\caption{Ablation study on the 
network designs. 
}
\vspace{-0.25cm}
\resizebox{0.35\textwidth}{!}{%
\begin{tabular}{|l|c|c|c|}
\hline
 Model &  Skip-connection & Offset-Attention & $L2$~$CD$ \\
\hline

A  &   &   & 1.15\\
B  & \checkmark   &   & 1.02 \\

C &   & \checkmark & 0.98  \\
D & \checkmark  & \checkmark &  0.92 \\\hline
\end{tabular}}
\label{exp:tab:abl_arc}
\end{table}

\begin{table}[t!]
\centering
\caption{Ablation study on 
the normalisation technique effect in  $L2$~$CD$ loss}
\resizebox{0.7\columnwidth}{!}{%
\vspace{-0.25cm}
\begin{tabular}{|l|c|c|}
\hline
 Method&  Baseline norm. & Our norm.\\
\hline

PCN  & 2.59  &  2.07  \\
 
PoinTr   & 1.66 &  1.15  \\
Ours     &   \textbf{1.44} & \textbf{0.92}  \\ \hline
\end{tabular}
}
\label{exp:tab:abl_alignment}
\end{table}

\subsubsection{\textbf{Point Cloud Generation}}
The main objective of our proposed PCD completion network is to predict the missing point cloud representing the unseen part of the object. To do that, we feed the predicted feature vector $AD$ (i.e., the output of the decoder) to the $FC$ layer, followed by Max-pooling and another $FC$ layer. Furthermore, the output of the last $FC$ layer will be concatenated with the predicted sparse point cloud $PS$  reconstructed by query $Q$, and pass through another $FC$ layer. Then, we utilise  FoldingNet \cite{yang2018foldingnet} $FN$ which is able to output a high resolution PCD by applying Fold operation on the output predicted feature vector of the missing PCD from decoder. We can define the point cloud generation process as: $PM =  FN(AD) + PS$, (symbol '$+$' represents set concatenation)  where $PM$ is the prediction of the missing parts of the point cloud. The predicted  missing point cloud $PM$ will be merged with the partial input point cloud $PP$ to shape the final complete point cloud $PC$ where $PC = \left\{ PC^i \in \mathbb{R}^3, i=1 \dots Z\right\}_{Z{= 8192}}$. We also fed the output feature vector of each encoder layer (by an element-wise summation) to the corresponding decoder layer using Skip-Connections (see Table \ref{exp:tab:abl_arc} for our design choice). 

\begin{table*}[t]

\caption{REAL ROBOT EXPERIMENT RESULT.}
\vspace{-0.25cm}
\centering
\resizebox{1\textwidth}{!}{%
\begin{tabular}{|l|c|c|c|c|c|c|c|c|c|c|c|}
\hline
 Method & Avg&  Pringles & Drill box &  Mustard & Mug & Cleanser &  Clamps (biggest) & Drill & Jell-o & Baseball & Pitcher with lid \\ 
\hline
GPD \cite{ten2017grasp}  & $46\%$ & $50\%$ & $0\%$ & $50\%$ & $70\%$ & $60\%$ & $30\%$ & $40\%$ & $80\%$ & $60\%$ & $20\%$ \\ 
\hline
GPD + ours  & $76\%$ & $80\%$&  $80\%$  &  $80\%$  & $80\%$ & $70\%$ & $70\%$ & $80\%$ & $90\%$ & $90\%$ & $40\%$ \\
\hline
\end{tabular}}
\label{exp:tab:grasp_sec_rate}
\end{table*}

\subsubsection{Network training}
\label{sec:method:Network_train}
Training is achieved by summation of  the  \emph{Chamfer-Distance (CD)} loss between the sparse and completed point cloud and   ground-truth point cloud:



\begin{equation}
\label{eq:loss}
\begin{aligned}
   L = \mathcal L_{cd} \left( PS, PGT \right) + \nonumber  
    \mathcal L_{cd} \left( PC, PGT\right),
\end{aligned}
\end{equation}
where $\mathcal L_{cd}$ is the \emph{Chamfer-Distance} loss \cite{yuan2018pcn}, $PGT$ is the ground-truth PCD, $PC$  is the the completed PCD, and $PS$ is the predicted sparse PCD. We supervise  both $PS$ and $PC$ using ground-truth completed point cloud during training to force both encoder and decoder about the complete shape of the GT PCD.

\subsection{Grasp pose generation}
\label{sec:method:Grasp pose generation and evaluation}
To generate the 6DoF grasp pose candidate for the two-fingered gripper, we use the Grasp Pose Detection (GPD) network introduced in \cite{ten2017grasp}. GPD  uniformly samples points in a Region of Interest (ROI) at random. ROIs are selected from an image-based object detection algorithm, but the algorithm can be tailored to the application's constraints. On the randomly sampled points of the ROIs, a local search heuristic is applied to find suitable orientations in the vicinity of each point, so a grasp candidate $GK$ corresponds to the sampled point and selected orientation. Then, the candidates $GK$ are classified as graspable by a four-layer CNN. The input of the CNN is a multiple view representation of the (clipped by the gripper) point cloud. To obtain the views, the PCD is voxelized, and the voxels are projected onto orthogonal axes. Finally, the grasps are ranked according to the output of the last layer of the CNN before the application of the Soft-max function. Thus, the classification score $CS$ provides the ranking of $GK$. In real-world experiments, grasps are executed according to their ranks. Each grasp candidate corresponds to the goal pose of the end-effector of the robotic arm. We use MoveIt to 
compute a collision-free trajectory for the arm to reach the target pose.

\section{EXPERIMENTS}
\label{exp:main}

The experiments are divided into two parts. In Section  \ref{sec:exp:comp}, we first evaluate the performance of our proposed PCD completion network against a range of state-of-the-art methods on the partial YCB dataset \cite{varley2017shape}. We also perform extensive ablation studies to justify the design choices of our PCD completion network architecture in terms of the Offset-Attention encoder-decoder and the skip-connection. We then present real robotic experiments in Section \ref{sec:exp:roboticgrasping} utilising a Kinova robotic arm equipped with an RGB-D sensor. We evaluate the grasp success rate of \mname~(i.e., GPD with our PCD completion network) in comparison to the one using only partial PCD. A grasp is considered successful if the object is held~(i.e. does not fall) after lifting it up. 





\begin{figure}[t!]
   \centering
   \includegraphics[width=0.5\textwidth]{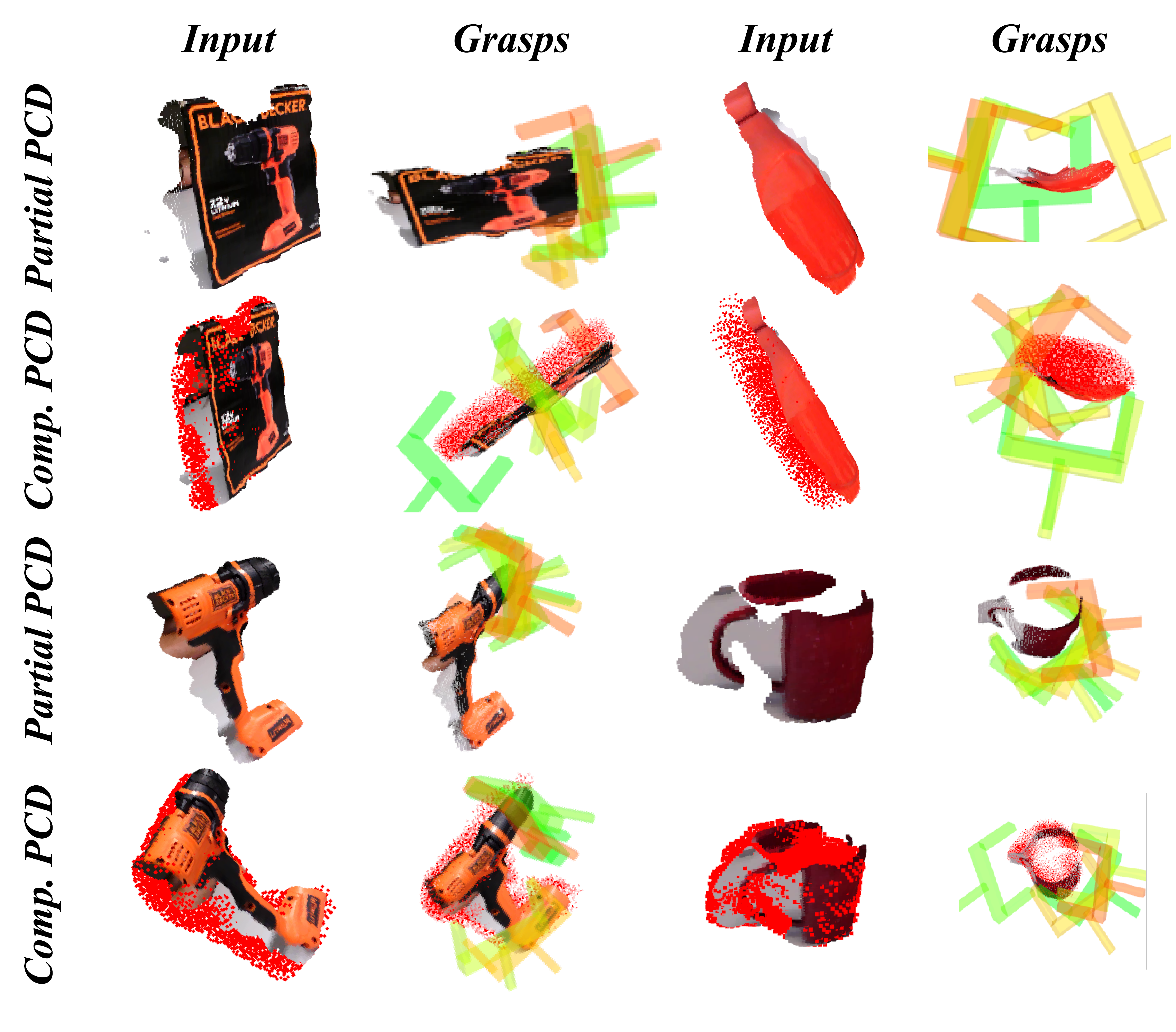}
    \vspace{-0.25cm}
   \caption{Qualitative results of generated grasp proposal on the top of partial and completed PCD of 4 objects using our PCD completion network. The partial PCDs are acquired by the real sensor on the Kinova robotic arm. Each candidate grasp pose generated by GPD is color-coded with green to red representing the score from high to low.}
    \label{fig:rcon}
\end{figure}

\subsection{Evaluation on 3D shape completion}
\label{sec:exp:comp}
\noindent\textbf{Dataset.}
We use the partial version of the YCB dataset \cite{varley2017shape}, which is a popular choice in PCD completion for robotic grasping \cite{lundell2019robust}. We randomly sample 50 views from the training set (Training Views), 50 views from the holdout view set (Holdout Views), and 50 views from the holdout models set (Holdout Models). 
We evaluate the completion network on holdout views and holdout model sets and the training is achieved with only the training set split.
As the exact train/test split is unspecified in \cite{varley2017shape}, for fair comparison with the state-of-the-art methods on PCD shape completion, we train all compared methods using our own split dataset. All PCDs are pre-processed as described in Section \ref{sec:method:alignment} for normalisation and sampling. 
\\
\noindent\textbf{Comparison.}
We compare our proposed PCD completion network against a range of state-of-the-art methods for shape completion in terms of the $L2$ \emph{Chamfer-Distance} loss \cite{yuan2018pcn} (multiplied by 1000)  between the reconstructed, and the ground truth PCD.  
For a fair comparison, we train from scratch (all the mentioned methods using the same dataset and split)  and test against the existing PCD completion networks such as FoldingNet \cite{yang2018foldingnet}, PCN \cite{yuan2018pcn}, MSN \cite{liu2020morphing}, and PoinTr \cite{yu2021PoinTr} on the partial YCB dataset using their open-source code with their best hyper-parameters. We are unable to fairly compare against \cite{chen2022improving} as the code is unavailable. 
As shown in Table~\ref{exp:tab:comp}, on average, our completion network achieves the lowest reconstruction loss among the competitors, outperforming the state-of-the-art method (i.e., PoinTr)
by $+0.23$. 


\noindent\textbf{Ablation.}
\label{exp:abl_arc_}
We perform extensive experiments to justify our network design choices in terms of the Offset-attention and skip-connection using the partial YCB dataset. Moreover, we evaluate the effect of our proposed PCD alignment processing technique on PCD completion performance.


\noindent\textbf{Do the Offset-attention and skip-connection improve the PCD completion accuracy?}
We evaluate the impact of our proposed Offset-Attention encoder-decoder layer and skip connection on the PCD reconstruction error. A set of variant models are studied: model $A$ is the baseline Transformer with Self-Attention encoder-decoder layer, model $B$ adds Skip-connection between the encoder and decoder to the baseline model, model $C$ replaces Self-Attention with Offset-Attention layer and model D adds both skip-connection and Offset-Attention layer to the transformer.
As shown in Table \ref{exp:tab:abl_arc}, we observed that using skip-connection can improve the performance of the baseline model by $ +0.13$. When using Offset-Attention layer (model $C$) instead of Offset-Attention layer (model $A$), we observe an improvement to $0.98$. The best result is achieved by model $D$ when adding both skip-connection and Offset-Attention layer to the transformer.

\noindent\textbf{Does the PCD pre-processing help with PCD completion?} We evaluate the effect of our proposed PCD pre-processing technique on point cloud completion using our completion network, PCN \cite{yuan2018pcn} and PoinTr \cite{yu2021PoinTr}. The $Baseline~ norm.$ stands for normalising the GT and partial PCD with same formula but different parameters and $Our~norm.$ use the parameters of partial PCD to normalise GT PCD (See Section \ref{sec:method:alignment}). As shown in Table \ref{exp:tab:abl_alignment}, our network achieved the lowest reconstruction error compared to the other two methods using both PCD processing techniques. Moreover, there is a large reduction in PCD reconstruction error when applying our proposed pre-processing technique to all methods.

\subsection{Evaluation on robotic grasping}
\label{sec:exp:roboticgrasping}
We perform the real-world experiments utilising a Kinova Gen3 robot equipped with a Robotiq 2F-85 gripper
for grasping and an Intel RealSense D430 depth camera to capture the point cloud. 
During the experiments, an object is placed on the table and the robot starts at a predefined initial pose facing the object as shown in Figure \ref{fig:first_ARC}. For each test, the partial PCD of the object is extracted by removing the background information using a PCD segmentation network \cite{qi2017pointnet++}. Then, the segmented point cloud is fed to our completion network. Finally, the GPD \cite{ten2017grasp} network generates and ranks grasp candidates for both the completed and the partial PCD. The final grasp is chosen as the best ranked and with a feasible solution when sent to the MoveIt! motion planner. 


Each object is grasped 10 times at different poses w.r.t. the arm in its workspace. We compare GPD vs. GPD with \mname in Table \ref{exp:tab:grasp_sec_rate}, where GPD with \mname consistently outperform GPD without object PCD completion. 
For example, the drill box dimensions (specially width, See figure \ref{fig:rcon}) are not captured by the partial PCD, which results in a low success rate due to collisions with the object. With our method, the completed PCD can better address this issue after correctly reconstructing its shape. 
Another hard case is the Pitcher with a lid given the size of the object and the gripper's maximum aperture. Since the lid reduces the available grasp poses from the top, only grasps from the handle are feasible. Nevertheless, our \mname doubles the successful grasps compared to the baseline with partial PCD.     


\section{CONCLUSIONS}

In this work, we propose a new system called \mname for improving the robotic grasp success rate in a real-world experiment. The central core of the proposed system is the PCD completion head with the ability to complete accurately the missing geometry of the 3D objects that have not been observed before and  without moving the camera to extract more information. We also proposed a new way to normalize partial views of PCD, solving the misalignment problem that improves robotic grasp success rate and reduces PCD completion error. With the experiment, we show that our network achieves a state-of-the-art result on PCD completion tasks and improves the average grasp success rate by a large margin. In future work, we will extend our work to multi-object shape completion and grasping. Moreover, we will investigate the possibility of a fusion of single-view RGB image and 3D PCD within the framework of \mname, to boost the 3D completion accuracy and grasp success rate.





\bibliographystyle{ieeetr}
\bibliography{IEEEabrv,refs}

\end{document}